\documentclass[review]{elsarticle}

\usepackage{hyperref}
\usepackage{tabularx}
\usepackage{amsmath}
\usepackage{booktabs} 
\usepackage{caption} 
\usepackage{adjustbox}
\usepackage{afterpage}
\usepackage{rotating}
\usepackage{soul}
\usepackage[table]{xcolor} 
\usepackage{changepage,threeparttable} 
\usepackage{subcaption}
\usepackage{caption}

\journal{Journal of \LaTeX\ Templates}

\begin{document}

\begin{frontmatter}

\title{Protein Secondary Structure Prediction Using 3D Graphs and Relation-Aware Message Passing Transformers}

\tnotetext[eqcontrib]{Equal contribution.}

\author[1]{Disha Varshney\tnotemark[eqcontrib]}
\author[2]{Samarth Garg\tnotemark[eqcontrib]}
\author[3]{Sarthak Tyagi}
\author[5]{Deeksha Varshney}
\author[4]{Nayan Deep}
\author[6]{Asif Ekbal}

\address[1]{Banaras Hindu University, Varanasi, India}
\address[2]{ABV-IIITM, Gwalior, India}
\address[3]{Indiana University, Bloomington}
\address[4]{IIIT, Naya Raipur, India}
\address[5]{Indian Institute of Technology Jodhpur, India}
\address[6]{Indian Institute of Technology Patna, India}

\ead{dishavarshney082@gmail.com, samarthgarg0904@gmail.com, tyagi.sarthakst2000@gmail.com, deeksha@iitj.ac.in, varshneynayan02@gmail.com, asif.ekbal@gmail.com}

\begin{abstract}
In this study, we tackle the challenging task of predicting secondary structures from protein primary sequences, a pivotal initial stride towards predicting tertiary structures, while yielding crucial insights into protein activity, relationships, and functions. Existing methods often utilize extensive sets of unlabeled amino acid sequences. However, these approaches neither explicitly capture nor harness the accessible protein 3D structural data, which is recognized as a decisive factor in dictating protein functions. To address this, we utilize protein residue graphs and introduce various forms of sequential or structural connections to capture enhanced spatial information. We adeptly combine Graph Neural Networks (GNNs) and Language Models (LMs), specifically utilizing a pre-trained transformer-based protein language model to encode amino acid sequences and employing message-passing mechanisms like GCN and R-GCN to capture geometric characteristics of protein structures. Employing convolution within a specific node's nearby region, including relations, we stack multiple convolutional layers to efficiently learn combined insights from the protein's spatial graph, revealing intricate interconnections and dependencies in its structural arrangement. To assess our model's performance, we employed the training dataset provided by NetSurfP-2.0, which outlines secondary structure in 3-and 8-states. {Extensive experiments show that our proposed model, SSRGNet surpasses the baseline on f1-scores.} 
\end{abstract}


\end{frontmatter} 

\section{Introduction}
Proteins serve as essential components within cells and are involved in various applications, spanning from therapeutics to materials. They are composed of a sequence of amino acids that fold into distinct shapes. With the development of affordable sequencing technologies \cite{ma2015novor,ma2012novo}, a substantial number of novel protein sequences have been identified in recent times. However, annotating the functional properties of a newly discovered protein sequence is still a laborious and expensive process. Thus, there is a need for reliable and efficient computational methods to accurately predict and assign functions to proteins, thereby bridging the gap between sequence information and functional knowledge. The analysis of protein structure, particularly the tertiary structure, is highly significant for practical applications related to proteins, such as understanding their functions and designing drugs \cite{noble2004protein}. Currently, there are three primary methods employed for predicting the tertiary structure of proteins. These include X-ray crystallography and nuclear magnetic resonance (NMR) techniques \cite{vvuthrich1989protein}, cryo-electron microscopy (cryo-EM) based methods \cite{wang2015novo}, and computer-aided ab initio prediction \cite{mandell2009computer}. Computer-assisted protein tertiary structure prediction has gained significant attention because it offers convenience and shows superior performance compared to other methods hindered by time-consuming processes of X-ray
crystallography, sequence length limitations involved with magnetic resonance (NMR), or expensive equipment requirements for cryo-EM. In the case of ab initio tertiary structure prediction, the protein's secondary structure plays a crucial role as it provides essential information about the local patterns in the protein's structure. Thus, improving the accuracy of protein secondary structure prediction (PSSP) is vital for achieving more precise predictions of the overall protein structure.

PSSP (Protein Secondary Structure Prediction) involves assigning a secondary structure label (coil, alpha helix, beta-sheet) to each amino acid in a protein sequence, representing the local structure. This process is analogous to sequence labeling in natural language processing (NLP). Several advanced deep learning models \cite{zhou2014deep,wang2016protein,li2016protein} have achieved satisfactory performance in PSSP by utilizing high-quality PSSM (Positional-Specific Scoring Matrix) in conjunction with one-hot encoded amino acid sequences as evolutionary information. \cite{zhou2014deep} employed a deep convolutional network to capture the relationship between PSSM features and labels. \cite{wang2016protein} proposed an enhanced model by incorporating a conditional random field after the convolutional neural network (CNN) to efficiently capture sequential dependencies. \cite{sonderby2014protein} addressed the problem using a two-layer LSTM (Long Short-Term Memory), while \cite{li2016protein} enhanced the representation power of the model by introducing GRU (Gated Recurrent Unit) units after the convolutional layers. \cite{Wang_Wang_Xu_Wu_Zhao_Li_Wang_Huang_Cui_2021} introduces a PSSM-Distil framework for protein secondary structure prediction, enhancing low-quality PSSM accuracy by leveraging teacher networks, feature enhancement, distribution alignment, and auxiliary information from a pre-trained protein sequence language model.

\begin{figure}[t!]
    \centering    \includegraphics[width=1.0\textwidth]{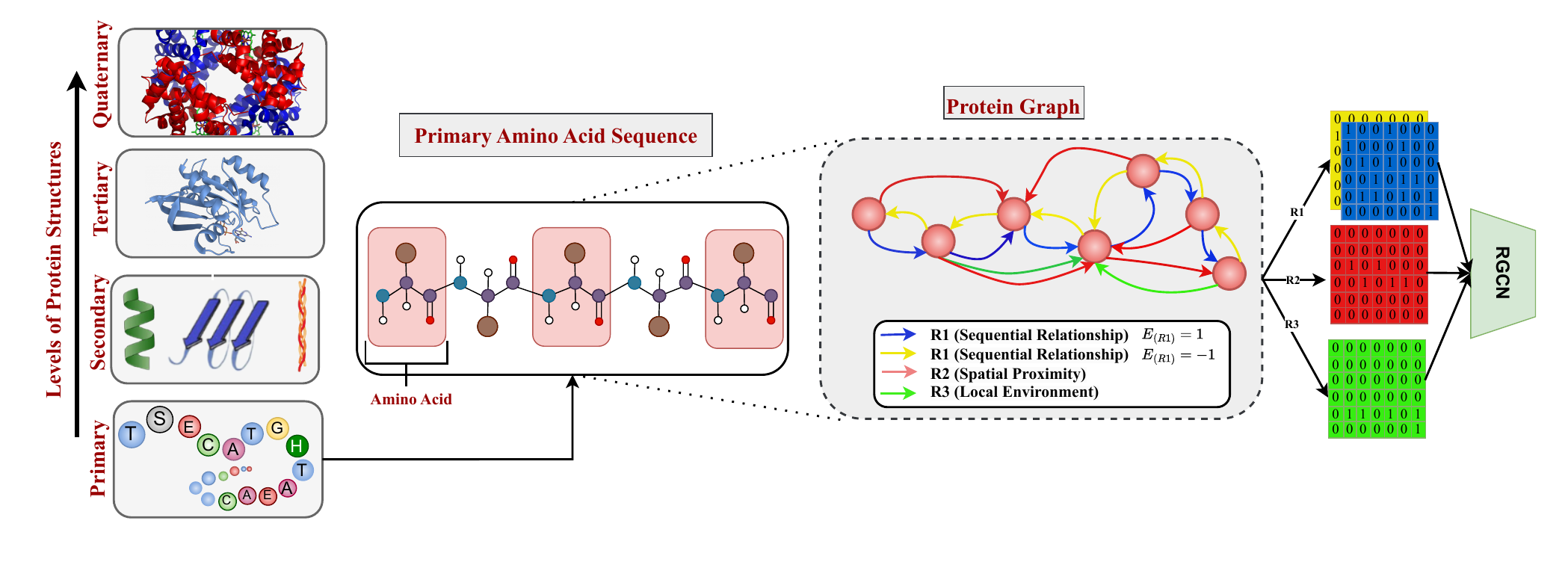}
    \caption{{Graph Representation of Protein Sequence: The figure illustrates the hierarchical levels of protein structures, ranging from the primary amino acid sequence to secondary, tertiary, and quaternary structures. The primary sequence is then transformed into a protein graph, where amino acids are represented as nodes and edges capture different relationships: R1 (sequential relationship) with $E_{R1}$=1 shown in \textcolor{blue}{blue} and with $E_{R1}$=-1 in \textcolor{yellow}{yellow}, R2 (spatial proximity) in  {red}, and R3 (local environment) in  \textcolor{green}{green}. This graph-based representation enables structured learning of protein properties using a Relational Graph Convolutional Network (RGCN).}}
    \label{fig:intro}
\end{figure}

Protein language models pre-trained on extensive protein sequence data have demonstrated impressive performance in various protein-related tasks, such as per-residue secondary structure prediction and per-protein localization and membrane prediction \cite{9477085, geffen2022distilprotbert}. However, while these models can implicitly capture inter-residue contact information \cite{doi:10.1073/pnas.2016239118}, they cannot explicitly encode protein structures to create structure-aware protein representations. To address this challenge, recent research has incorporated protein structures to determine secondary structure \cite{9733590,9669366}. Nevertheless, these models often overlook the vital interactions between edges in the protein structure, interactions that play a pivotal role in modeling protein structures \cite{jumper2021highly}. To bridge this gap and advance large-scale structural bioinformatics, the development of highly accurate computational methods that effectively utilize critical structural information is essential. This structural information encompasses details about the spatial arrangement of amino acids, the chemical bonds connecting them, and the overall architecture of the protein's three-dimensional structure. Understanding this data is paramount for unraveling the processes of protein folding and functioning, as a protein's biological activity is intricately linked to its specific three-dimensional structure. {Figure \ref{fig:intro} visually presents the graph structure corresponding to an amino acid sequence.} In this representation, a sequence of amino acids is depicted as a graph, with each amino acid residue represented as a node and the connections between these nodes as edges. Various types of edges, as elaborated in Section \ref{sec:graph_const}, play crucial roles in representing this structural information. This graph-based approach holds promise for capturing essential protein structure information.

Motivated by this development, we enhance the Protein Language Models with a structure-based encoder for protein secondary structure prediction. We propose a novel yet efficient architecture named Secondary Structure Relational Graph Neural Network (SSRGNet). This approach utilizes protein-residue graphs along with amino acid chains. In the protein-residue graphs, each node corresponds to an amino acid's alpha carbon and includes its 3D coordinates. We enhance these graphs by incorporating diverse edges to account for various sequential and structural relationships between residues. To encode sequence information, we employ a state-of-the-art Protein Language Model named \textit{DistilProtBert} \cite{geffen2022distilprotbert}. For structural encoding, we use a simple yet effective protein structure encoder which performs relational message passing on the enhanced protein graphs. This is the first attempt to integrate relational message passing in GNNs for protein structure encoding to predict secondary structures. {We investigate three model architectures that fuse the two encoders in a series, parallel, and cross manner.} We conduct experiments utilizing the NetSurfP-2.0 benchmark for protein secondary structure prediction. The results on CB513, TS115, and CASP12 test sets verify the superiority of the proposed SSRGNet over vanilla Protein Language Models, various protein encoders, and existing structure encoder-enhanced Protein Language Models. These results illustrate the great promise of relation-based structure-encoder-enhanced Protein Language Models for PSSP.


Our major contributions can be summarized as follows:

\begin{enumerate}
    \item We introduce a novel and efficient architecture named the \textit{Secondary Structure Relational Graph Neural Network (SSRGNet)} which adeptly facilitates protein secondary structure prediction by synergistically harnessing both sequence information and structural details through protein-residue graphs and amino acid chains.
    \item We enrich the protein-residue graphs by integrating diverse types of edges to encapsulate various relationships between residues, capturing sequential patterns, spatially close interactions,
    and the impact of the local environment on the protein’s structure.
    \item Through rigorous experimentation utilizing the NetSurfP-2.0 benchmark, we affirm the superiority of SSRGNet over baseline models on prominent test sets like CB513, TS115, and CASP12, thereby confirming the promising prospects of relation-based structure-encoder-enhanced protein language models for Protein Secondary Structure Prediction.
\end{enumerate}

\section{Related Works}
In this section, we explore past efforts that have employed various statistical methods, machine learning algorithms, and deep learning methodologies to address challenges in PSSP.

\subsection{Statistical Methods}
Initially, protein secondary structure prediction relied on statistical methods. These approaches were built upon statistical information regarding individual amino acids and were limited in their application to a small number of proteins with known structures. Despite their relatively basic nature, these methods still serve as an initial step in addressing the protein secondary structure problem. Two examples of such methods are the Chou-Fasman algorithm \cite{prevelige1989chou} and the Garnier-Osguthorpe-Robson algorithm \cite{garnier1989gor}. The Chou-Fasman algorithm \cite{prevelige1989chou} utilizes amino acid frequencies to determine the occurrence of helices, sheets, and coils based on X-ray crystallography data from known protein structures. By establishing probability parameters derived from these frequencies, the algorithm predicts the presence of helices, sheets, or coils within a given amino acid sequence in a protein. The prediction accuracy of the Chou-Fasman algorithm falls within the range of 50\% to 60\% \cite{kabsch1983good}. The Garnier-Osguthorpe-Robson (GOR) algorithm, similar to the Chou-Fasman algorithm, is a theory-based approach for protein secondary structure prediction. The GOR method utilizes probability parameters obtained from X-ray crystallography data of established protein structures. Yet, its approach isn't limited to these specific secondary structure probabilities. The method also takes into account the conditional probability of adjacent structures that share the same formation. Studies indicate that the GOR algorithm's prediction accuracy hovers around 57\%.

\subsection{Machine learning methods}
Over the years, numerous machine learning approaches, including Support Vector Machines (SVM), Dynamic Bayesian networks, Random Forests (RF), and ensemble methods, have been utilized to predict protein secondary structure. A frequently employed approach for predicting protein secondary structure is the Support Vector Machine (SVM). SVM typically utilizes a linear hyperplane to classify data. \cite{nguyen2003multi} applied SVM to a dataset generated from PSSM values and four physicochemical features, achieving a Q3 accuracy of 79.5\%.
\cite{shuai2017novel} introduced a novel radical group encoding approach for predicting protein secondary structure (PSSP). Their encoding methodology aims to represent the 20 common amino acids constituting a protein by utilizing information derived from the stable atomic structures present within these amino acids. The researchers conducted experiments on two datasets, CB513 and 25PDB. The models used for classification are support vector machines (SVM) and Bayes classifiers. Comparative analysis against the quadrature encoding method demonstrated superior performance, yielding an improved accuracy by 1.2\%.
Furthermore, researchers have used the Hidden Markov Model (HMM) for predicting protein secondary structures \cite{asai1993prediction}. HMM is a statistical model that predicts future behavior by analyzing existing data and is widely employed as a classifier in diverse fields, such as bioinformatics, data mining, pattern recognition, and image processing, among others. In the work by \cite{aydin2006protein}, an extended hidden semi-markov model was employed for single sequence secondary structure prediction, achieving a Q3 accuracy of 67.9\% on the CASP6 dataset.

Occasionally, classification algorithms may exhibit analogous errors when compared with one another, resulting in committing errors associated with specific classes. In order to mitigate these types of errors, ensemble methods use mathematical and statistical techniques to amalgamate two or more classification algorithms. An illustration of this approach can be found in the work of \cite{bouziane2015effect}, who combined artificial neural networks (ANN) and support vector machines (SVM) using the weighted pooling ensemble which assigns more weight to highly accurate classifiers, yielding a Q3 accuracy of 78.50\% for RS126 dataset and 76.65\% for CB513. \cite{jones1999protein} employed neural networks on the Position-Specific Scoring Matrix (PSSM) calculated using the PSI-BLAST algorithm, achieving a Q3 accuracy range of 76.5\% to 78.3\%. In a study conducted by \cite{ghosh2008protein}, k-NN, minimum distance, and fuzzy k-NN algorithms were applied to a dataset of protein structures, and their performance was compared against multilayer perceptron networks. The findings revealed that these methods outperformed multilayer perceptron networks, achieving higher accuracy. \cite{saha2013protein} proposed an ensemble technique employing two stochastic supervised machine learning algorithms, Maximum Entropy Markov Model (MEMM) and Conditional Random Field (CRF).

\subsection{Deep Learning-based Methods}
The main objective of protein secondary structure prediction is to classify individual amino acids into specific secondary structural elements, such as the alpha helix, beta sheet, and coil. Given that the dataset size aligns with the number of amino acids, which can be substantial, the acceleration of the learning algorithm becomes pivotal. The authors in \cite{holley1989protein} employed a feed-forward neural network to predict protein secondary structure. They evaluated this method using a dataset of 64 proteins, where the initial 48 proteins were allocated for training, and the subsequent 14 were reserved for testing. The achieved accuracy on this dataset was reported as 79\%. An innovative method that employs two-dimensional deep convolutional neural networks to address the task of managing extensive protein datasets was proposed in \cite{liu2017novel}. This network structure comprises six convolutional layers and five max-pooling layers, facilitating efficient feature extraction and pattern recognition. They evaluated the model's performance by conducting experiments on six benchmark datasets, namely 25PDB and CASP\{9–13\}. The DCRNN model \cite{li2016protein} utilizes multi-scale convolutional neural networks in conjunction with stacked bidirectional gated recurrent units to capture both local and broader contextual information. Furthermore, \cite{busia2017next} utilized deep convolutional neural networks and achieved an even higher Q8 accuracy of 71.4\% using the CB513 dataset. The work by \cite{zhou2014deep} presents a supervised generative stochastic network (GSN) and incorporates a convolutional architecture to enable hierarchical representation learning on full-sized data. It achieved 66.4\% Q8 accuracy on the CB513 dataset. A Bidirectional Recurrent Neural Network (BRNN) with Long Short-Term Memory (LSTM) units was employed for protein secondary structure prediction in \cite{sonderby2014protein}, resulting in a Q8 accuracy of 67.4\% on the CB513 dataset. A combination of recurrent neural networks and 2D convolutional neural networks proposed in \cite{guo2018protein}, achieved a Q8 accuracy of 70\%.  Additionally, \cite{kumar2020enhanced} utilized a deep learning framework with hybrid profile-based features and reported Q8 accuracy scores of 75.8\% and 73.5\% on the CB513 and CB6133 datasets, respectively.

SSpro8 \cite{magnan2014sspro}, employs an ensemble of 100 bidirectional recursive neural networks to predict an eight-state protein secondary structure. DeepCNF \cite{wang2015deepcnf}, melds the strengths of both conditional neural fields (CNF) and deep convolutional neural networks to not only encapsulate medium- and long-range sequence information but also articulate the interdependency of order/disorder labels among neighboring residues field to model. Their method was evaluated using the CASP9 and CASP10 datasets. The study \cite{wang2017protein} introduced the SSREDNs, a deep recurrent encoder-decoder network that leverages deep layers and recurrent mechanisms to discern the complex nonlinear associations between protein input features and their Secondary Structure (SS). Additionally, it effectively models the interplay between successive residues in the protein sequence. The model was trained using the widely recognized CullPDB dataset and evaluated on both CullPDB and CB513. DeepACLSTM \cite{guo2019deepaclstm} utilizes Asymmetric Convolutional Networks (ACNNs) in conjunction with bidirectional Long Short-Term Memory (LSTM) units for secondary structure prediction. ACNNs delve into the intricate local contexts surrounding amino acids, while the BLSTM neural networks grasp the extended interrelationships among amino acids. It is trained using the CB5534 and evaluated on 3 publicly available test datasets \textit{viz.} CB513, CASP10 and CASP11. In \cite{heffernan2017capturing}, an iterative learning strategy is used to train the LSTM-BRNN neural network which is capable of capturing long-range interactions without using a window. The model is trained on TR4590 and evaluated on the TS1199 dataset. MUFOLD-SS \cite{fang2018mufold} is an open-source standalone tool for PSSP. The approach employs a novel deep inception-inside-inception (Deep3I) architecture to capture interactions between amino acids, encompassing both local and distant relationships. They evaluated the model using CB513, CASP10, CASP11, and CASP12 benchmarks. \cite{yang2022protein} proposed a lightweight convolutional network using a label distribution aware margin loss which helps learn minority class representation for PSSP. They trained the model using CB12510 and evaluated using CB513, CASP13, CASP10,
CASP12, CASP11 and CASP14 as test sets. Moreover, many other deep network variants have also been proposed to perform eight-state prediction \cite{hanson2019improving,klausen2019netsurfp,zhou2018cnnh_pss,ismi2024self}. 

In recent years, language models based on protein sequences, termed Protein Language Models, have emerged as potent tools for various protein-related tasks. Pioneering works such as those reported in \cite{9477085, geffen2022distilprotbert} showcased the impressive performances of models pre-trained on extensive protein sequence data for tasks like per-residue secondary structure prediction, and per-protein localization and membrane prediction. These models are lauded for their ability to implicitly capture inter-residue contact information as highlighted in \cite{doi:10.1073/pnas.2016239118}. Utilizing masked language modeling losses, existing models \cite{rives2021biological,ferruz2022protgpt2} are adept at encapsulating co-evolutionary data and implicitly seizing inter-residue contact information. However, a notable limitation lies in their inability to explicitly encode protein structures which precludes the creation of structure-aware protein representations. This gap underscores a pivotal avenue for future research aimed at enhancing the structural encoding capabilities of protein models, thereby pushing the boundaries of what can be achieved in protein analysis and prediction tasks.  {MFTrans \cite{chen2024mftrans}, a deep learning-based multi-feature fusion network, employs an MSA Transformer combined with a hybrid CNN-BiGRU architecture to integrate sequence, evolutionary, and hidden state information, enhancing protein secondary structure prediction accuracy through a multi-view feature fusion strategy.}

In comparison to sequence-based approaches, structure-based methods are theoretically more adept at learning a rich protein representation, given that a protein's functionality is fundamentally dictated by its structure. This line of works seeks to encode spatial information in protein structures by 3D CNNs \cite{derevyanko2018deep} or graph neural networks (GNNs) \cite{gligorijevic2021structure, baldassarre2021graphqa, wang2022learning, jing2020learning}. For the PSSP task, \cite{kihara2005effect} discusses the effect of long-range residue interactions on defining secondary structure in a protein and introduces the Residue Contact Order (RCO) metric to study the relationship between residue separation and prediction accuracy, utilizing a dataset of 2777 non-homologous proteins. Utilizing their finding, the authors in \cite{9669366} employed Graph Convolutional Networks (GCN) to amalgamate the information pertaining to amino acids and their interactions. On the other hand, Bi-Long Short Term Memory (Bi-LSTM) was utilized owing to its robust capability to apprehend the long-range dependencies among amino acids. They leveraged ProtTrans \cite{9477085} for obtaining sequence embeddings, and utilized the filtered CB6133 dataset for training purposes. The evaluation of performance was carried out on various datasets including CASP10, CASP11, CASP12, CB513, and TS115. \cite{nahid2021protein} creates a graph from the primary amino acid sequence, with unique embeddings for each node via orthogonal encoding. Through iterative traversal and GNN methodology, information from neighboring nodes is aggregated. A Support Vector Machine (SVM) is then applied to identify the 8 states of protein secondary structure, achieving a promising 76.89\% accuracy on the ccPDB 2.0 dataset. IGPRED \cite{gormez2021igpred} combining a convolutional neural network and graph convolutional network, utilizes features like PSI-BLAST PSSM, HHMAKE PSSM, and amino acids' physicochemical properties, refined through Bayesian optimization. It shows notable Q3 accuracies on different datasets including CullPDB, EVAset, CASP10, CASP11, and CASP12, indicating its effectiveness in protein secondary structure prediction.

In this work, the focus is on predicting secondary structures from protein primary sequences, which serves as a crucial step toward tertiary structure prediction, shedding light on protein functions and relationships. While prevailing methods rely on unlabeled amino acid sequences, they often overlook valuable protein 3D structural data, a key determinant of protein functions. To bridge this gap, we leverage the protein residue graphs, incorporating various types of relations between the edges to capture spatial information. These graphs are further refined by integrating various types of edges to encapsulate different interactions among residues, sequential trends, spatial proximity interactions, and the influence of the local milieu on the protein’s form. We conducted extensive experiments using the NetSurfP-2.0 benchmark and evaluated our model on significant test datasets like CB513, TS115, and CASP12, thereby showing the encouraging potential of relation-oriented structure-encoder-augmented protein language models for Protein Secondary Structure Prediction.

\section{Methodology}
\subsection{Problem Statement}
\label{sec:prob}
\begin{figure*}[t!]
    \centering    \includegraphics[width=0.90\textwidth]{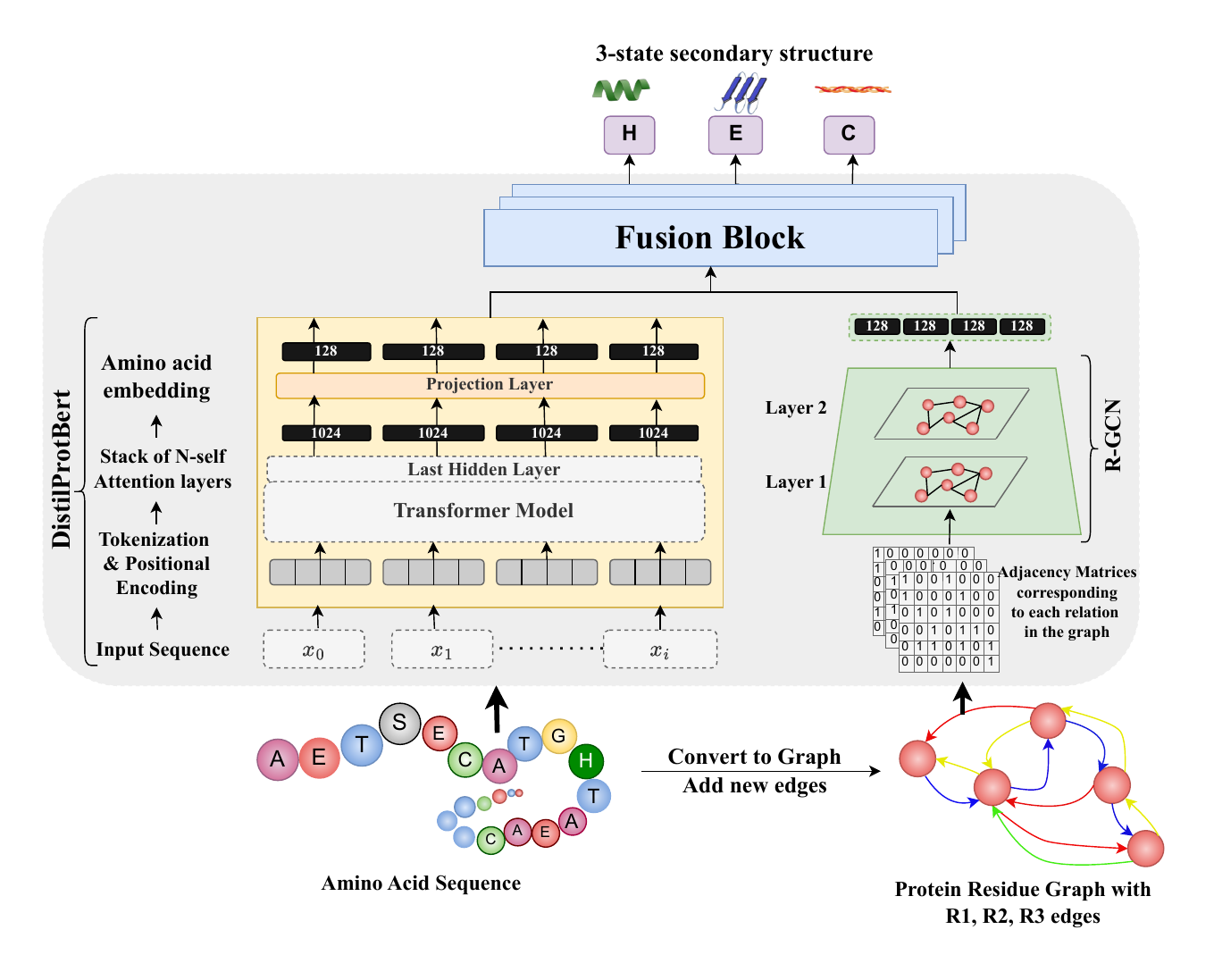}
    \caption{SSRGNet Model Overview: The diagram illustrates the SSRGNet model, which combines a DistilProtBert model, and an R-GCN (Relational Graph Convolution Network) model. The amino acid sequence is converted into a graph representation featuring three types of edges, as shown at the bottom, and passed into the respective encoders. Features obtained for the amino acid and the protein graph are finally fused together to predict the three-state and eight-state secondary structure of the protein.}
    \label{fig:method}
\end{figure*}

The protein sequence is composed of 21 standard amino acids \textit{viz.} \textit{Ala (A), Arg (R), Asn (N), Asp (D), Cys (C), Gln (Q), Glu (E), Glx (Z), Gly (G), His (H), Lle (I), Leu (L), Lys (K), Met (M), Phe (F), Pro (P), Ser (S), Thr (T), Trp (W), Tyr (Y), Val (V)}, which serve as its fundamental components. Additionally, the infrequent amino acids "U, Z, O, B" have been substituted with the symbol "X." We frame PSSP as a sequence-to-sequence task where each amino acid $x_i$ in a protein sequence $P$ is mapped to a label $y_i$ $\in$ \{alpha-helix (H), beta-strand (E), Coil (C)\} for 3-state classification and $y_i$ $\in$ \{residue in isolated $\beta$-bridge (B), Extended strand(E), 3-10 helix (G), alpha-helix (H), $\pi$-helix (I), Hydrogen bonded turn (T), Bend (S), Loop or any other residues (C)\} for 8-state classification. Figure \ref{fig:method} provides a detailed overview of our proposed methodology.

\subsection{Sequence Encoding}
\label{sec:sequence_enc}
Protein sequences can provide valuable information about the likelihood of a specific amino acid's involvement in forming a secondary structure. This is due to the distinct electrochemical and geometric characteristics exhibited by each amino acid. To encode the amino acid sequence $x_i$, we use DistilProtBert \cite{geffen2022distilprotbert} as our base model. {It is a distilled version of the ProtBert-UniRef100 model. DistilProtBert is trained on UniRef100, a comprehensive dataset containing clustered protein sequences. Apart from using techniques like cross entropy and cosine teacher-student losses, DistilProtBert was trained using a masked language modeling (MLM) approach,  where random amino acids are masked, the model learns to predict them based on contextual residues and it exclusively processes amino acids in uppercase letters. DistilProtBert is primarily a protein language model that employs a transformer-based architecture and attention mechanism to understand how different pairs of amino acids interact within a sequence. The protein sequence is tokenized and positional encoding is incorporated. The ensuing vectors are channeled through the DistilProtBert model to generate context-aware embeddings for each input token, i.e., each amino acid. Here, we utilized the last hidden state of the Transformer’s attention stack as input for the downstream protein secondary prediction task. This enables the DistilProtBert model to grasp the evolutionary details of proteins and their characteristics. } 

In order to achieve uniform representation dimensions for all modalities within the spatial domain, a projection layer is applied after the last hidden layer of the DistilProtBert model. This layer functions as a projection layer that transforms the individual amino-acid representations, derived from the DistilProtBert embedding dimension, $d_p \in 1024$, into the graph embedding dimension $d_g$. As a result, a matrix denoted as $H^{d_g \in 128}$ is formed, containing the amino-acid representations. 

\subsection{Graph Construction}
\label{sec:graph_const}
In this work, we start by visualizing the 3D structure of proteins as a graph. Within this graph, every amino acid residue is portrayed as an individual node, and the relations between them as an edge. Emphasis is placed on including different types of relationships among the edges to encapsulate a variety of spatial information. In the case of modeling protein interactions using a multi-relational heterogeneous graph, we start by defining the graph \(G = (V, E)\) as follows:

\begin{enumerate}
    \item \(V = \{x_i\}\) is the set of all nodes, where:
    \begin{itemize}
        \item \(x_i\) is the set of amino acid residues, representing the individual units that form proteins. 
    \end{itemize}
    
    \item \(E\) is the set of edges. All edges \((u, v, r) \in E\) have a source \(u\), a target \(v\), and a relation type \(r \in R\). The set of possible relations \(R = \{R1, R2, R3\}\) contains:
    
    \begin{itemize}
        \item \(R1\) (Sequential Relationship):
        \begin{itemize}
            \item An edge is established between two amino acid residues if the number of amino acids separating them in the protein sequence is less than a specified threshold value (\(d_s\)).
            \item The type of edge is determined based on the relative position of the residues in the sequence, resulting in $E_{R1}$ = \(2d_s - 1\) distinctive types of edges that capture their local relationships.
            \item We choose \(d_s = 2\) as the threshold value. This choice results in three types of edges, each representing a different relationship between amino acid residues.
        \end{itemize}
        
        \item \(R2\) (Spatial Proximity):
        \begin{itemize}
            \item An edge is added between two amino acid residues if their Euclidean distance in the three-dimensional structure of the protein falls below a specified threshold value (\(d_{ed}\)).
            \item We choose $(d_{ed} = 10 $\AA$)$ as the threshold value \cite{zhang2022protein}. This means that residues that are physically close to each other in the protein's three-dimensional space and are within 10 \AA, will be connected by an edge.
        \end{itemize}
        
        \item \(R3\) (Local Environment):
        \begin{itemize}
            \item For each amino acid residue node, its \(k\) closest neighbors in three-dimensional space are identified based on the Euclidean distance.
            \item These nearby nodes are then connected to the target residue node, allowing the graph to incorporate the impact of the local neighborhood.
            \item The parameter \(k_n\) controls the number of nearest neighbors considered, adjusting the extent of local connectivity within the graph. In this case, \(k_n = 10\) for the experiments.
        \end{itemize}
    \end{itemize}
\end{enumerate}

\subsection{Graph Features}
\label{sec:graph_feat}
In the resulting graph, every amino acid residue is portrayed as an individual node, and the characteristics of these residues are encapsulated in the node's features. Additionally, the connections between residues are translated into edges, and the attributes of these edges capture the diverse types of spatial interactions between the amino acids. Both nodes and edges within the protein graph are enriched with features that stem from the protein's sequence and structure. These features are calculated based on the specific attributes of the protein, offering a comprehensive representation of its composition and arrangement.

\begin{enumerate}
\item Node Features: As we explicitly depict the protein's 3D structure in a graph, each node solely contains the sequence features corresponding to the respective residue. The identity and conservation of a residue are quantified by 21 features that capture the relative frequency of each of the 21 amino acids in alignment with similar proteins. The encoding process of the amino acid sequences is similar to the encoding of protein sequences as explained in Section \ref{sec:sequence_enc}. The resulting representation matrix, $H^{d_g}$, is the node feature matrix.

\item Edge Features: Alongside these sequence-based attributes, each node incorporates features derived from the protein's structure. These encompass the residue's local context, its sequential patterns, spatially close interactions, and the influence of the local environment \cite{zhang2022protein} as explained in Section \ref{sec:graph_const}. The edge features are encoded using the method described in Section \ref{sec:method_rgcn}.
\end{enumerate}

\subsection{Methods for Relation-based Graph Convolution}
\label{sec:method_rgcn}
Using the protein graphs and their features, we aim to model representations that encode their spatial and chemical information. To meet this requirement, we start by constructing our protein graph based on spatial features as detailed in Section \ref{sec:graph_const}. This construction of a multi-relational heterogeneous graph allows us to delve into the spatial relationships between amino acid residues and grasp the impact of the local environment on protein structure.

Using the protein graphs described earlier, we apply a Graph Neural Network (GNN) to obtain representations at both the individual amino acid and complete protein levels. One straightforward example of a GNN is the Graph Convolutional Network (GCN) \cite{kipf2016semi}, where messages are generated by multiplying node features with a convolutional kernel matrix that is shared across all edges. For the given heterogeneous graph, we employ a two-layer R-GCN (\cite{schlichtkrull2018modeling}), a graph neural network designed specifically for learning representations in multi-relational data. The R-GCN updates the initial node embeddings in each layer by considering the neighborhood of each node and taking into account the type of relation it has with its neighbors. In other words, for every node \(x \in V\), the R-GCN calculates its embedding, \(e^{(k+1)}_x\), in the $(k+1)$-th convolutional layer as follows:

\begin{equation}
e^{(k+1)}_x = \sigma \left( \frac{1}{{c_{x,r}}} \sum_{r \in R} \sum_{j \in N_r(x)} W^{(k)}_r e^{(k)}_j + W^{(k)}_0 e^{(k)}_x \right)
\end{equation}

\(N_r(x)\) refers to the set of neighbors of node \(x\) connected by relations of type \(r\). The activation function, \(\sigma\), normalization constant, \(c_{x,r}\), and relation-specific transformations, \(W_r\) and \(W_0\) are incorporated by the R-GCN during training. Following the approach proposed by \cite{schlichtkrull2018modeling}, we set \(c_{x,r} = |N_r(x)|\): the number of neighbors of node \( x \) connected by relation \( r \), enabling our R-GCN to generate comprehensive representations of proteins based on their interactions, spatial proximity, and local environment. Utilizing convolution in a specified node's neighboring area, inclusive of relations, we layer multiple convolutional stages to effectively extract consolidated insights from the protein's spatial graph, unveiling complex interlinks and dependencies in its structural configuration.

\subsection{Fusion Block}
We aim to efficiently combine graph neural networks and pretrained language models, particularly employing a pre-trained transformer-based protein language model to encode amino acid sequences, and utilizing message-passing mechanisms within networks to grasp the geometric attributes of protein structures. The sequence features ($S_{\text{seq}}$) from Section \ref{sec:graph_const} and graph features ($S_{\text{graph}}$) from Section \ref{sec:graph_feat} are concatenated together along the last dimension.

\begin{gather}
    F = \text{MLP}(\text{concat}(S_{\text{seq}}, S_{\text{graph}})) \\
    p_{i,j} = \frac{e^{F_i}}{\sum_{j=1}^N e^{F_j}}
\end{gather}

where, 
$S_{\text{seq}} = H$, 
$S_{\text{graph}} = e$.

\subsection{Training}
For the three-state protein secondary structure prediction, the cross-entropy loss can be defined as:
\begin{equation}
\mathcal{L} = -\sum_{i=1}^{N}\sum_{j=1}^{C} y_{ij} \log(p_{ij})
\end{equation}
where, 
\begin{itemize}

\item \( N \) is the number of samples,
\item \( C \) is the number of classes (e.g., helix, sheet, coil),
\item \( y_{ij} \) is the true label, and
\item \( p_{ij} \) is the predicted probability of sample \( i \) belonging to class \( j \).
\end{itemize}

\section{Experiments}
In this section, we assess how well the proposed method works for Protein Secondary Structure Prediction (PSSP). We start by introducing the datasets used and explaining the evaluation metrics used. Then, we detail the implementation and report the experimental results under various conditions.

\subsection{Datasets}
We utilized the NetSurfP-2.0\footnote{https://services.healthtech.dtu.dk/services/NetSurfP-2.0/training\_data/Train\_HHblits.npz}\cite{klausen2019netsurfp} dataset, which characterizes secondary structure in 3-and 8-states, to predict attributes associated with individual tokens, specifically single amino acids, referred to as residues when they form part of proteins. Initially, the structural dataset comprised 12,185 crystal structures sourced from the Protein Data Bank (PDB)\cite{berman2000protein} and filtered by the PISCES server with specific criteria, including a 25\% sequence similarity clustering threshold and a resolution of 2.5 Å or better. To prevent overfitting, they removed sequences that shared more than 25\% identity with any sequences in the test datasets, as well as peptide chains containing fewer than 20 residues, resulting in 10,837 sequences. Subsequently, they randomly selected 500 sequences (allocated as the test set) for early stopping and parameter optimization, leaving 10,337 sequences for training. Therefore, the training and test sets comprised 10,337 non-redundant proteins and 500 sequences, respectively. To evaluate the performance of the proposed methods, we also use three commonly used datasets, the CB513\footnote{{https://services.healthtech.dtu.dk/services/NetSurfP-2.0/training\_data/CB513\_HHblits.npz}} (513 protein regions from 434 proteins) \cite{yang2018sixty}, TS115\footnote{https://services.healthtech.dtu.dk/services/NetSurfP-2.0/training\_data/TS115\_HHblits.npz} (115 proteins) \cite{cuff1999evaluation}, and CASP12\footnote{https://services.healthtech.dtu.dk/services/NetSurfP-2.0/training\_data/CASP12\_HHblits.npz} (21 proteins) \cite{abriata2018assessment}. {Table \ref{tab:table1} and Table \ref{tab:table2} shows a detailed overview of the 3-state and 8-state class distribution secondary structure respectively for the NetSurfP-2.0 training set and the corresponding validation datasets (CASP12, TS115, CB513)\cite{klausen2019netsurfp}.} In all the datasets we've used, the Loop or other residue is denoted by "C" instead of "L" for eight-state classes. This notation has been consistently applied in our paper.

\textbf{3D-coordinates of Amino Acids:} We have used the PDB IDs linked with the samples in the dataset to fetch the PDB files from the PDB data bank\footnote{https://www.rcsb.org/}. We extract the coordinates of each of the residues in the long protein sequence and use them for constructing protein graphs as detailed in Section \ref{sec:graph_const}. For some PDB IDs in the NetSurfP-2.0 dataset, we could not retrieve the corresponding PDB files. The statistics mentioned in Table \ref{tab:table1} and Table \ref{tab:table2} are reported considering the previous factor.

\begin{table}[ht!]
  \centering
  \caption{ \label{tab:table1} Class distribution (3-state secondary structure) for the NetSurfP-2.0 training set and the corresponding validation datasets (CASP12, TS115, CB513).}
  \begin{tabular}{|c|c|c|c|c|}
    \hline
    \textbf{Dataset} & \textbf{Protein} & \textbf{Helix (H)} & \textbf{Strand (E)} & \textbf{Coil (C)} \\
    \hline
    CASP12 & 21 & 1478 & 2241 & 2701 \\
    TS115 & 115 & 5380 & 11641 & 10480 \\
    CB513 & 511 & 18690 & 29102 & 35635 \\
    NetSurfP-2.0 & 10796 & 585603 & 1173670 & 1001075 \\
    \hline
  \end{tabular}

\end{table}

\begin{table}[ht!]
  \centering

\renewcommand{\arraystretch}{1.2}
\setlength\tabcolsep{1.2pt}
    \caption{\label{tab:table2} Class distribution (8-state secondary structure) for the NetSurfP-2.0 training set and the corresponding validation datasets (CASP12, TS115, CB513).}
      \begin{adjustbox}{max width=1.0\textwidth}
    \begin{tabular}{|c|c|c|c|c|c|c|c|c|}
      \hline
      \textbf{Dataset} & \textbf{H} & \textbf{E} & \textbf{C} & \textbf{S} & \textbf{T} & \textbf{G} & \textbf{B} & \textbf{I} \\
      \hline
      CASP12 & 1989 & 1416 & 1400 & 668 & 633 & 215 & 62 & 37 \\
      TS115 & 10434 & 5085 & 5395 & 2210 & 2875 & 1033 & 295 & 174 \\
      CB513 & 25559 & 17585 & 17713 & 8211 & 9711 & 3074 & 1105 & 469 \\
      NetSurfP-2.0 & 887210 & 559154 & 681602 & 209690 & 282378 & 99697 & 26409 & 14168 \\
      \hline
    \end{tabular}%
  \end{adjustbox}
\end{table}

\subsection{Implementation Details}
\label{sec:imp_det}

\subsubsection{Language Model}
The protein sequences were converted to uppercase and tokenized using a single space, with a vocabulary size of 21. Pre-processing for longer sequences exceeding 1024 amino acids was done. All the experiments are implemented using the Pytorch framework \cite{paszke2019pytorch}. The last hidden layer size for DistilProtBert model is 1024, and the number of layers is 15. We use the AdamW \cite{loshchilov2017decoupled} optimizer with a learning rate fixed to 1e-05. 

\subsubsection{Graph Model}
We opted for PyG\footnote{https://pytorch-geometric.readthedocs.io/en/latest/} due to its extensive collection of state-of-the-art models for implementing graph neural networks. However, a challenge arose because PyG models don't naturally handle batch inputs, while the hugging face Language Model (LM) outputs data in batches. To overcome this, we utilized diagonal batching, a PyG technique that enables parallelization across multiple samples. In this approach, we stack adjacency matrices diagonally, creating a large network with isolated subgraphs. Node and target attributes are concatenated in the node dimension. To manage data batching effectively, we developed a custom PyG in-memory dataset object. R-GCN and GCN models had a hidden size of 128, and the number of layers was set to 2. We again use the AdamW \cite{loshchilov2017decoupled} optimizer with a learning rate fixed to 3e-04. 
\\ \\
Our model required distinct inputs for language and graph components. LM needed protein sequences, while the graph model required an adjacency matrix to represent residue relationships, along with a node feature matrix generated by LM. Initially, this matrix was absent, so we dynamically passed it to the graph model during the forward call of the training process.

For hyperparameter optimization, we employed a robust Bayesian search strategy to identify the optimal learning rates (lr) for both the Protein Language Model and the Graph Neural Networks. Using validation loss as the evaluation metric, we fine-tuned hyperparameters to pinpoint the most suitable learning rate (lr) values. Bayesian Optimization was pivotal in creating a probabilistic model that maps hyperparameter values to their corresponding target values from the validation set. This systematic exploration of the hyperparameter space led to the selection of learning rate (lr) values that significantly improved the performance of our SSRGNet model, ultimately enhancing the accuracy and efficiency of protein secondary structure predictions. We choose the best model when the loss on the validation set does not decrease further. We use the GeForce GTX 2080 Ti as the computing infrastructure. Our codes are available at this link\footnote{ \href{https://github.com/SamarthGarg09/protein-secondary-structure-prediction}{https://github.com/SamarthGarg09/protein-secondary-structure-prediction}}.

\subsubsection{Time Complexity Analysis}

\paragraph{BERT-base Model Time Complexity}

The BERT (Bidirectional Encoder Representations from Transformers) model has a computational complexity for each of its 12 layers as follows:

\begin{itemize}
    \item \textbf{Self-Attention Mechanism}: $O(N^2 \cdot d)$
    \item \textbf{Feed-Forward Network}: $O(N \cdot d^2)$
\end{itemize}

Thus, the overall complexity for BERT-base is:
\[ O(N^2 \cdot d + N \cdot d^2) \]

\paragraph{RGCN Model Time Complexity}

For a Relational Graph Convolutional Network (RGCN) with $K$ layers:

\begin{itemize}
    \item \textbf{Message Passing}: $O(|E| \cdot d)$
    \item \textbf{Update Step}: $O(|V| \cdot d^2)$
\end{itemize}

The total complexity is:
\[ O(K \cdot (|E| \cdot d + |V| \cdot d^2)) \]

\paragraph{Combined Model Complexity}

Combining BERT and RGCN results in:

\begin{itemize}
    \item \textbf{BERT}: $O(N^2 \cdot d + N \cdot d^2)$
    \item \textbf{RGCN}: $O(K \cdot (|E| \cdot d + |V| \cdot d^2))$
\end{itemize}

The combined complexity is:
\[ O(N^2 \cdot d + N \cdot d^2 + K \cdot (|E| \cdot d + |V| \cdot d^2)) \]

This provides insight into the computational demands of our model, aiding optimization for large datasets and complex protein structures.

\subsection{Evaluation metrics}
The efficacy of the suggested approach and the baseline is assessed through two criteria: Accuracy, F1-score.

\subsubsection{Accuracy}
The Q3 accuracy metric determines the percentage of amino acid residues whose predicted secondary structure aligns with the actual structure \cite{ma2018generating}. Its calculation is articulated in Eq. (8):
\[
Q3 = \frac{N_H + N_E + N_C}{N} \times 100 \quad (8)
\]
Here, $N_H$ signifies the correctly forecasted helix, $N_E$ the right strand prediction, $N_C$ the accurate coil prediction, and $N$ encompasses the total amino acid residues in a given protein. 

Similarly, the Q8 accuracy metric is computed as delineated in Eq. (9):
{\footnotesize
\[
Q8 = \frac{N_E + N_H + N_S + N_T + N_B + N_C + N_G + N_I}{N} \times 100 \quad (9)
\]
}
Herein, \( N_C \) corresponds to the right prediction of Loop or any other residues, \( N_B \) is the correctly anticipated \( \beta \)-bridge, \( N_E \) is the precise extended strand prediction, \(N_H\) is the correct prediction of alpha-helix, \(N_I\) is $\pi$-helix , \(N_T\) is Hydrogen bonded turn, \(N_G\) is 3-10 helix and \(N_S\) is Bend while \( N \) stands for the cumulative count of amino acid residues in a protein.

\subsubsection{F1-Score}
The F1-Score represents the harmonic average of precision and recall. 

\paragraph{\textbf{Precision}}
Precision is the proportion of accurate positive predictions to the total number of positive predictions made. This metric can be mathematically represented as shown in Eq. (5):
\[
\text{Precision} = \frac{TP}{TP + FP} \quad (5)
\]
Here, True Positive (TP) signifies the instances where the predicted positive outcome is indeed positive. On the other hand, False Positive (FP) represents the cases where a negative outcome is incorrectly classified as positive.

\paragraph{\textbf{Recall}}
The proportion of accurate positive predictions relative to all predicted instances within the specified class is termed as recall, as depicted in Eq. (6):
\[
\text{Recall} = \frac{TP}{TP + FN} \quad (6)
\]
where FN stands for the false negative value, signifying a positive value that has been erroneously predicted as negative.

The metric achieves its maximum value of 1 when both recall and precision are perfect. Conversely, if either recall or precision registers a value of zero, the F1-Score subsequently becomes zero. The formula to calculate the F1-Score is provided in Eq. (7):
\[
F1 = \frac{2 \times \text{precision} \times \text{recall}}{\text{precision} + \text{recall}} \quad (7)
\]

\subsection{Baseline models}
In our experimental evaluation, we conducted a rigorous performance assessment of the SSRGNet framework by comparing it to a comprehensive set of baseline models. Initially, we evaluated our approach against unimodal encoders, including DCRNN \cite{li2016protein}, DeepACLSTM \cite{guo2019deepaclstm}, DistilProtBert \cite{geffen2022distilprotbert}. These encoders exclusively focus on representing protein sequences. Furthermore, we compared our proposed model to a multimodal baseline named GCNBLSTM \cite{9669366} that concatenates the graph and sequence representations without fusing the representation of each amino acid and the structural representation.

\begin{enumerate}

\item  \textbf{DCRNN}  \cite{li2016protein}: The Deep Convolutional and Recurrent Neural Network (DCRNN) comprises of a feature embedding layer, multi-scale CNN layers, three BiGRU layers, and two fully connected hidden layers. It processes protein amino acid sequences by transforming sparse sequence features into denser vectors and extracting both local and global contextual features through various layers.

\item  \textbf{DeepACLSTM}  \cite{guo2019deepaclstm}: DeepACLSTM combines protein sequence and profile attributes through asymmetric convolutional neural networks (ACNNs) and bidirectional long short-term memory (BiLSTM) systems. The ACNNs decipher the complex local nuances of amino acids, while the BLSTM systems identify the extensive interconnections among them, ensuring each amino acid residue is categorized considering both its immediate environment and wider interdependencies.

\item \textbf{DistilProtBert} \cite{geffen2022distilprotbert}: DistilProtBert is a distilled version of the ProtBert model, designed for bioinformatics tasks, particularly in protein sequence analysis. It leverages the power of the BERT architecture, trained on a massive corpus of protein sequences, to understand the contextual relationships between amino acids in a given protein sequence.

\item \textbf{SSGNet}: SSGNet combines the DistilProtBert model and a GCN (Graph Convolution Network) model to provide structural insights using the protein residue graph (Section \ref{sec:graph_const}) for the secondary structure prediction task.

\end{enumerate}

\subsection{Ablation Models}
In this section, we present an ablation study to investigate the impact of different fusion techniques, namely Series Fusion, Cross Fusion, and Parallel Fusion, on the performance of our model (Figure \ref{fig:fusion}). Each fusion technique represents a unique methodology for processing and combining information. We employed three distinct fusion methodologies explained as follows:

\begin{figure*}[h!]
    \centering    \includegraphics[width=1.0\textwidth]{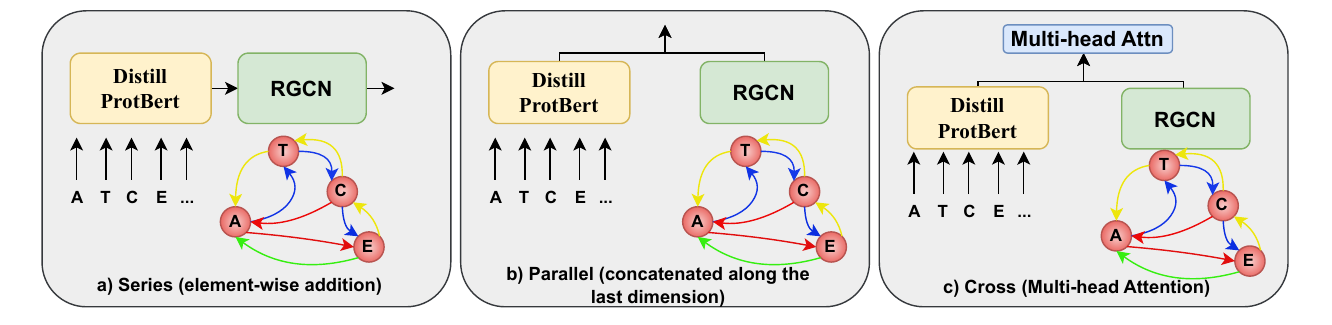}
    \caption{Comparative Architectural Schemes of Fusion Techniques. The diagram illustrates three distinct methodologies: (a) Series Fusion, where components process information sequentially (element-wise addition); (b) Parallel Fusion, where outputs are concatenated along the last dimension, and (c) Cross Fusion, characterized by intertwined (multi-head attention) processing layers.}
    \label{fig:fusion}
\end{figure*}

\begin{itemize}
    \item \textbf{Series Fusion}: This technique processes information the sequence encoder (Section \ref{sec:sequence_enc}) and graph encoder (Section \ref{sec:method_rgcn}) sequentially through element-wise addition.
    \item \textbf{Parallel Fusion}: In this the outputs from the sequence encoder and graph encoder are concatenated along the last dimension (Eq. (2)).
    \item \textbf{Cross Fusion}: This method of fusion is characterized by intertwined processing layers employing multi-head attention mechanisms.
\end{itemize}


\section{Results and Analysis}
We present the outcomes in Table \ref{tab:3-state_res} and Table \ref{tab:8-state_res},  involving various encoder models, encompassing unimodal encoders such as DCRNN \cite{li2016protein}, DeepACLSTM \cite{guo2019deepaclstm}, DistilProtBert \cite{geffen2022distilprotbert}, and our multimodal baseline named SSGNet. The unimodal DCRNN and DeepACLSTM baseline, which relies solely on the amino acid sequence of the protein, exhibits the lowest performance across all evaluation metrics for both three-state and eight-state class predictions. The metric of accuracy can be misleading in imbalanced datasets as it might predominantly reflect the predictions of the majority class. On the other hand, the F1 score offers a balanced insight into a model's performance by considering both precision and recall, making it a more reliable metric in scenarios where false positives and false negatives have differing implications. Despite this, we conduct a comprehensive analysis employing both metrics. A marked improvement in performance is observed upon utilizing the pre-trained protein language model. {DistilProtBert significantly outperforms the DCRNN and DeepACLSTM, improving by over 41\% on the test set of the NetSurfP-2.0 dataset when evaluated using the F1 score. This substantial enhancement can be ascribed to the pretraining regimen of DistilProtBert, which employs masked protein modeling. This approach enables the encoder to discern intricate relationships and patterns within protein sequences, thereby contributing to the observed performance boost.}

\begin{table*}[!ht]
  \centering
  \caption{\label{tab:3-state_res}
{Comparison of results for predictions in three states in test sets of the NetSurfP-2.0, CB513, CASP12, and TS115 datasets. The best performance for F1 is highlighted in bold and improvement over the best baseline, and is statistically significant (t-test with p-value at 0.05 significance level).}}
\begin{center}
\begin{adjustbox}{max width=1.0\textwidth}
\renewcommand{\arraystretch}{1.0}
\setlength\tabcolsep{1.3pt}

  \scriptsize
  \begin{tabular}{|>{\centering\arraybackslash}p{1.9cm}|cc|cc|cc|cc|}
    \toprule
& \multicolumn{2}{c|}{\textbf{NetSurfP-2.0}} & \multicolumn{2}{c|}{\textbf{CB513}} & \multicolumn{2}{c|}{\textbf{CASP12}} & \multicolumn{2}{c|}{\textbf{TS115}} \\
\cmidrule(r){2-3} \cmidrule(lr){4-5} \cmidrule(lr){6-7} \cmidrule(l){8-9}
& Accuracy (\%) & F1 & Accuracy (\%) & F1 & Accuracy (\%) & F1 & Accuracy (\%) & F1 \\
    \cmidrule{1-9} 
DCRNN  & 20.66 & 0.46 & 95.47 & 0.44 & 15.59 & 0.44 & 16.91 & 0.46 \\
DeepACLSTM & 95.16 & 0.47 & 95.19 & 0.45 & 93.58 & 0.44 & 95.52 & 0.48 \\
DistilProtBert & 80.60 & 0.65 & 79.49 & 0.61 & 69.74 & 0.50 & 80.93 & 0.65 \\
\cmidrule{1-9}
\textbf{SSRGNet} & 81.23 & \textbf{0.67} & 80.17 & {0.61} & 71.93 & \textbf{0.51} & 82.01 & {0.65} \\
SSGNet  & 81.10 & 0.66 & 80.02 & \textbf{0.62} & 71.79 & \textbf{0.51} & 81.74 & \textbf{0.66} \\
    \bottomrule
  \end{tabular}
  \end{adjustbox}
\end{center}
\end{table*}

\begin{table*}[!ht]
\centering
\caption{\label{tab:8-state_res}{Comparison of results for eight-state predictions on the test sets of NetSurfP-2.0, CB513, CASP12, and TS115 datasets. The top performance for F1 is denoted in bold and improvement over the best baseline, and is statistically significant (t-test with p-value at 0.05 significance level).}}
\begin{center}
\begin{adjustbox}{max width=1.0\textwidth}
\renewcommand{\arraystretch}{1.0}
\setlength\tabcolsep{1.3pt}
\scriptsize
\begin{tabular}{|>{\centering\arraybackslash}p{1.9cm}|cc|cc|cc|cc|}
    \toprule
& \multicolumn{2}{c|}{\textbf{NetSurfP-2.0}} & \multicolumn{2}{c|}{\textbf{CB513}} & \multicolumn{2}{c|}{\textbf{CASP12}} & \multicolumn{2}{c|}{\textbf{TS115}} \\
\cmidrule(r){2-3} \cmidrule(lr){4-5} \cmidrule(lr){6-7} \cmidrule(l){8-9}
& Accuracy (\%) & F1 & Accuracy (\%) & F1 & Accuracy (\%) & F1 & Accuracy (\%) & F1 \\
\cmidrule{1-9}
DCRNN & 15.47 & 0.50 & 16.40 & 0.48 & 12.85 & 0.46 & 14.29 & 0.48 \\
DeepACLSTM & 15.88 & 0.47 & 17.07 & 0.45 & 12.87 & 0.45 & 14.24 & 0.46 \\
DistilProtBert & 69.50 & 0.62 & 65.56 & \textbf{0.58} & 58.78 & \textbf{0.51} & 70.21 & 0.61 \\
\cmidrule{1-9}
\textbf{SSRGNet} & 70.23 & \textbf{0.63} & 65.67 & 0.57 & 58.61 & {0.50} & 70.33 & 0.60 \\
SSGNet & 69.93 & 0.62 & 66.06 & \textbf{0.58} & 59.27 & 0.50 & 71.14 & \textbf{0.62} \\
\bottomrule
\end{tabular}
\end{adjustbox}
\end{center}
\end{table*}

A notable improvement in performance is witnessed when the unimodal graph encoder GCN is employed within SSGNet, allowing it to surpass the performance benchmarks set by DCRNN and DeepACLSTM. Our proposed frameworks, SSRGNet and SSGNet—both with and without relations—exhibit superior performance over baseline models, including the DistilProtBert model across all test sets. {This marked difference in performance accentuates the importance of incorporating structural insights through the SSRGNet encoder for more accurate protein secondary structure prediction.} This endeavor underscores the importance of enriching the protein residue graph, as it facilitates a more nuanced understanding and representation of the intricate relationships and interactions that govern protein structure. By doing so, it enables the development of more robust and accurate models for protein secondary structure prediction, ultimately contributing to advancing the state of the art in this domain. Through this enriched representation, our approach leverages a holistic understanding of the multi-faceted interactions at play, thereby significantly enhancing the predictive capabilities of the SSRGNet framework in discerning the protein's secondary structure.

{The limited improvement in F1 score (~1\%) compared to DistilProtBert alone can be attributed to the dominance of sequence-based features in DistilProtBert, which already captures essential secondary structure information, reducing the additional benefit of structural encoding. The R-GCN’s contribution may be constrained by noisy or incomplete PDB data, limited edge definitions in the protein residue graph, or an inability to capture long-range dependencies effectively. Additionally, the fusion strategy might not optimally integrate sequence and structural features, leading to redundancy rather than complementarity. The data imbalance, particularly for underrepresented classes like $\pi$-helices and isolated $\beta$-bridges, further limits the structural encoder’s impact. Also, since that our experimental setup utilized a batch size of 1 during training. This choice was primarily dictated by memory and computational constraints given the model complexity and dataset size. Consequently, the reported performance metrics might not fully reflect the potential of our approach. We believe that training with a larger batch size would likely lead to improved generalization and more competitive scores.}

\begin{figure*}[t!]
     \centering
     \begin{subfigure}[b]{0.48\textwidth}
         \centering
         \includegraphics[width=\textwidth]{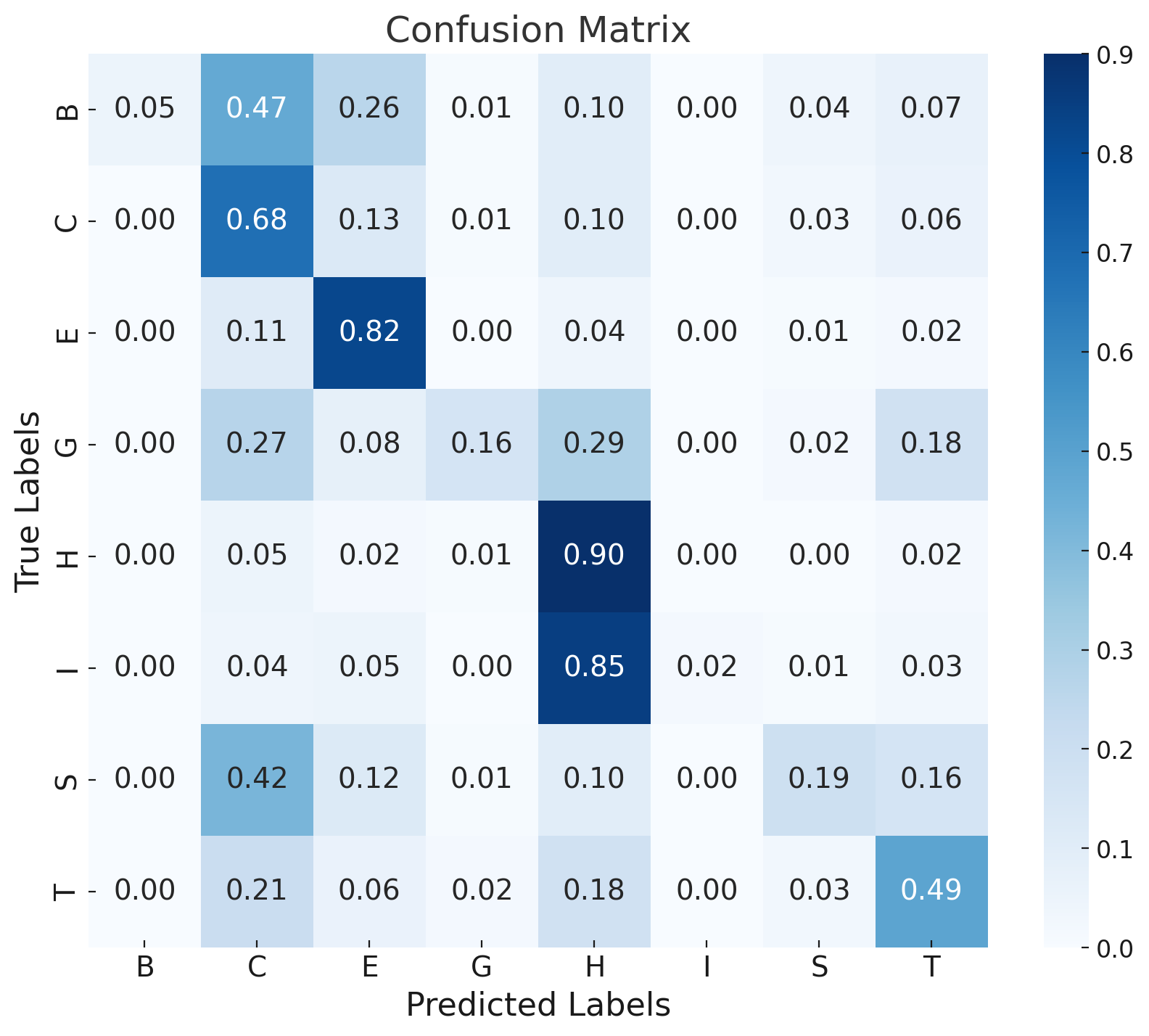}
         \caption{{NetSurfP-2.0}}
         \label{fig:net_surf}
     \end{subfigure}
     \hfill
     \begin{subfigure}[b]{0.48\textwidth}
         \centering
         \includegraphics[width=\textwidth]{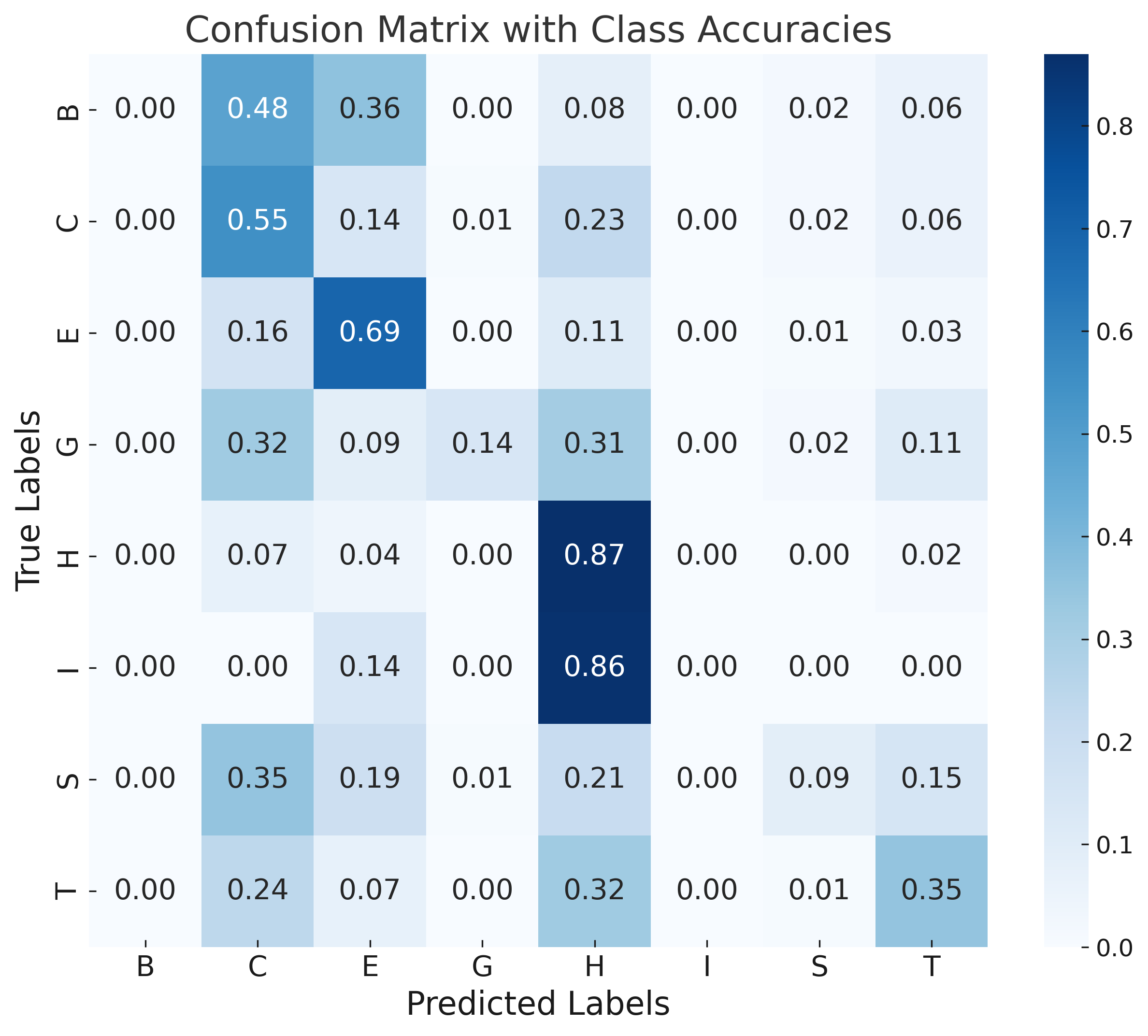}
         \caption{{CASP12}}
         \label{fig:casp12}
     \end{subfigure}
     
     \vspace{1em} 
     
     \begin{subfigure}[b]{0.48\textwidth}
         \centering
         \includegraphics[width=\textwidth]{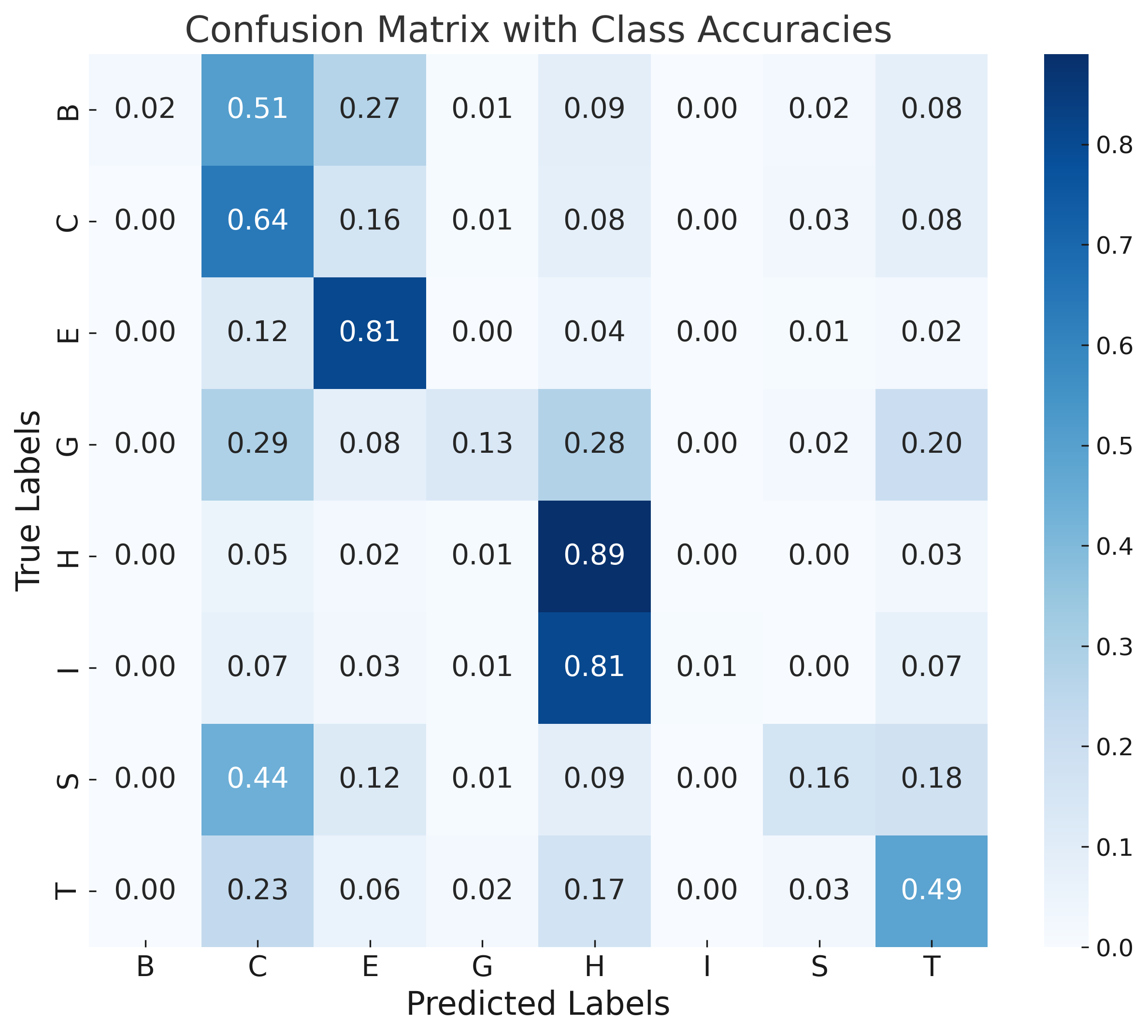}
         \caption{{CB513}}
         \label{fig:cb513}
     \end{subfigure}
     \hfill
     \begin{subfigure}[b]{0.48\textwidth}
         \centering
         \includegraphics[width=\textwidth]{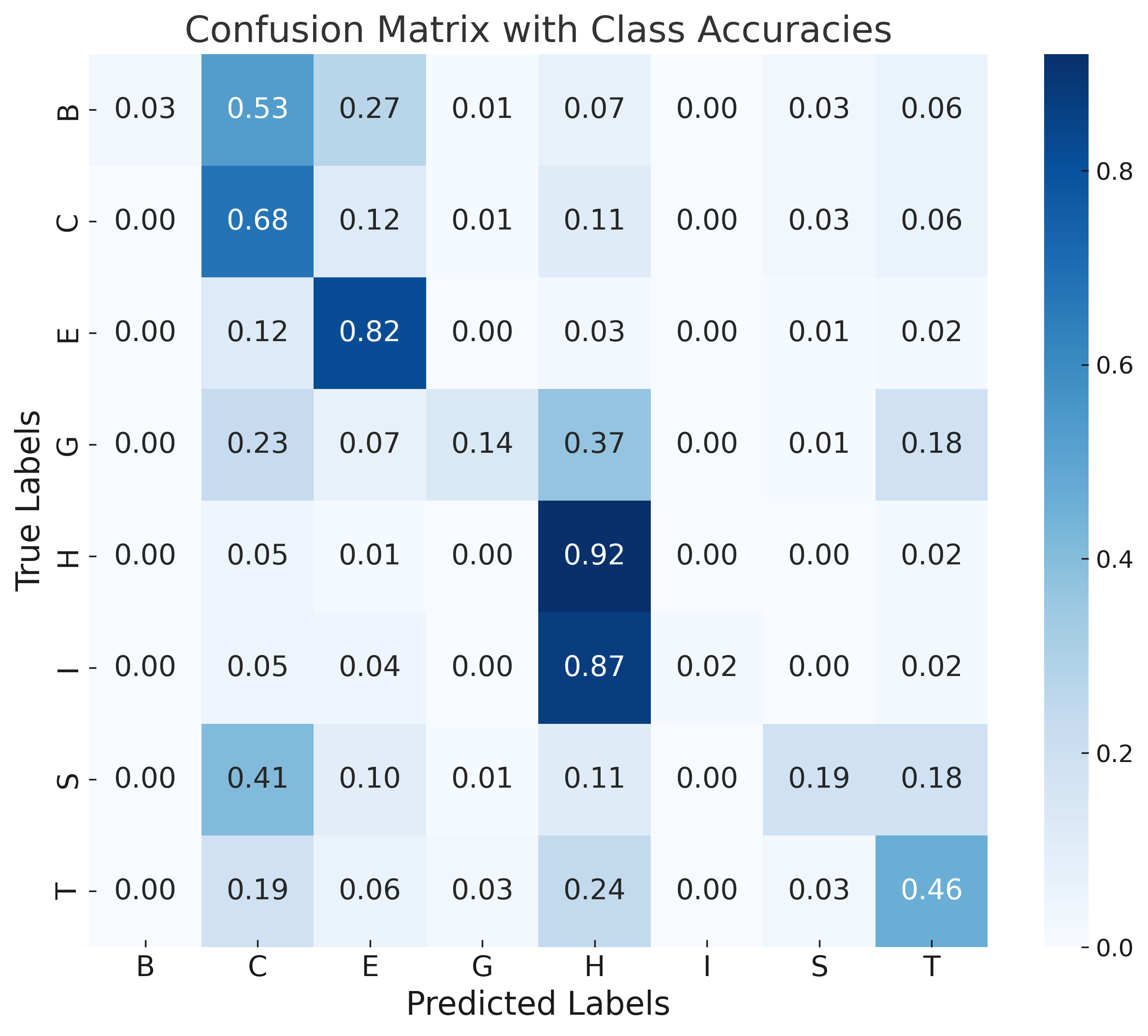}
         \caption{{TS115}}
         \label{fig:ts115}
     \end{subfigure}
     
     \caption{Confusion matrices for SSRGNet on different datasets for eight-state prediction.}
     \label{fig:confusion_matrices}
\end{figure*}

The confusion matrices presented in Figures \{\ref{fig:net_surf}, \ref{fig:casp12}, \ref{fig:cb513}, \ref{fig:ts115}\} illustrate the results of SSRGNet on the test sets of NetSurfP-2.0, CB513, CASP12, and TS115. Each entry in these matrices denotes how frequently a particular true category is misclassified into another category. One can observe that the \(\pi\)-helix (I), which is the least common class, is predominantly misclassified as an \(\alpha\)-helix (H). The isolated \(\beta\) bridge (B), another rare class, is primarily misidentified as either an irregular structure (L) or an extended strand (E). Furthermore, the 3-10 helix (G) is often mistaken for an irregular structure (L) or an \(\alpha\)-helix (H) across all datasets. Lastly, the Bend (S) class is most commonly misclassified as an irregular structure (L). {It's evident that more attention needs to be directed toward enhancing the classification accuracy of these underrepresented classes in subsequent research efforts. These misclassifications likely arise due to the scarcity of training examples for \(\pi\)-helices, isolated \(\beta\)-bridges, and 3-10 helices in the dataset, making it difficult for the model to learn clear decision boundaries for these classes. Additionally, their structural similarities to more frequent classes like \(\alpha\)-helices and extended strands lead to higher confusion.}

\begin{figure*}[t!]
     \begin{subfigure}[b]{0.50\textwidth}
         \centering
         \includegraphics[width=\textwidth]{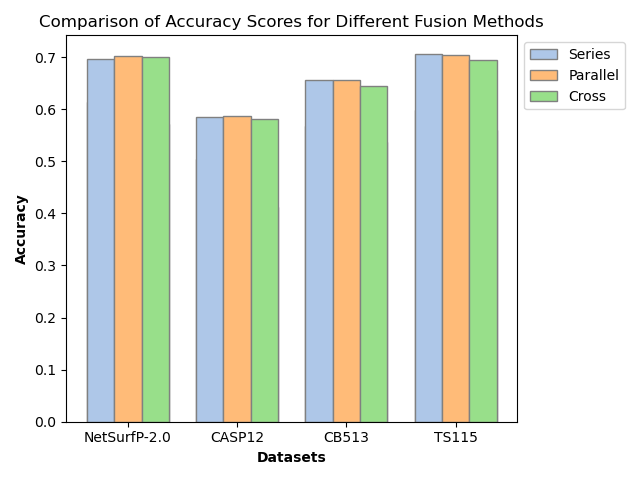}
         \caption{{Accuracy}}
         \label{fig:accuracy}
     \end{subfigure}
     \hfill
     \begin{subfigure}[b]{0.50\textwidth}
         \centering
         \includegraphics[width=\textwidth]{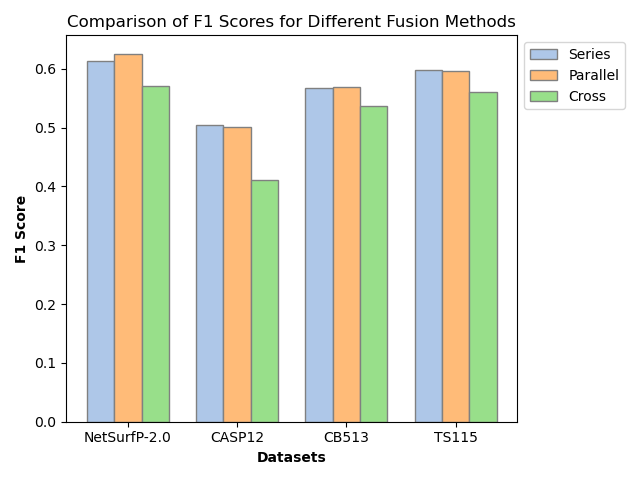}
         \caption{{F1-scores}}
         \label{fig:f1_scores}
     \end{subfigure}
     
     \caption{{Results for the Ablation Study on different evaluation metrics. Comparison of Loss, Accuracy and F1-scores for different fusion methods on various datasets.}}
     \label{fig:ablation_study}
\end{figure*}

\textbf{Ablation Study} An ablation study was conducted to assess the performance of our SSRGNet framework across different fusion techniques. The primary objective of this experiment was to evaluate the benefits of employing various fusion strategies in enhancing the performance of the SSRGNet framework. To conduct the ablation study, we systematically varied the fusion techniques while keeping other parameters constant, which allowed us to isolate and understand the impact of each fusion method on the overall performance of the framework. We evaluated each configuration on the same test set of proteins (\textit{viz.}  the test set of NetSurfP-2.0, CB513, CASP12, TS115 dataset) and used the same evaluation metrics as described earlier. The results of the ablation study, presented in Figure \ref{fig:accuracy}, \ref{fig:f1_scores}\}, show a trend of performance as we use different fusion methods. 
First, upon comparing Accuracy values, it is evident that parallel fusion, while simple in concept, is remarkably effective, surpassing the other two fusion strategies for all the test sets. Similarly, for F1-scores, we observe a similar pattern , where parallel fusion improves over serial and cross fusion on majority of the test set. Due to this pattern, we propose parallel fusion as our final strategy for encoding sequential and structural information of proteins. {Parallel fusion likely performs better because it preserves both sequence and structural information without forcing the model to prioritize one over the other. Serial fusion may lead to information loss as it combines features in a sequential manner, while cross fusion introduces additional complexity without significant performance gains. The ability of parallel fusion to retain complementary information from both modalities likely contributes to its effectiveness.}

\begin{table*}[!htp]
\centering
\small
\caption{\label{tab:ablation_results_2}{Comparison of ablation study results for eight-state predictions on the CB513, CASP12, and TS115 datasets. Full SSRGNet denotes the full fusion model, while R1 (Sequential), R2 (Spatial Proximity), and R3 (Local Environment) denote models using only the corresponding relation.}}
\begin{center}
\begin{adjustbox}{max width=1.0\textwidth}
\renewcommand{\arraystretch}{1.0}
\setlength\tabcolsep{1.3pt}
\scriptsize
\begin{tabular}{|>{\centering\arraybackslash}p{3.9cm}|cc|cc|cc|}
\toprule
 & \multicolumn{2}{c|}{\textbf{CB513}} & \multicolumn{2}{c|}{\textbf{CASP12}} & \multicolumn{2}{c|}{\textbf{TS115}} \\
\cmidrule(lr){2-3} \cmidrule(lr){4-5} \cmidrule(l){6-7}
Model Variant & Accuracy (\%) & F1 & Accuracy (\%) & F1 & Accuracy (\%) & F1 \\
\midrule
SSRGNet    & 65.67& 0.57 & 58.62 & 0.50 & 70.33 & 0.60 \\
R1 (Sequential Relationship) & 65.61 & 0.56 & 58.60 & 0.49 & 70.30 & 0.58 \\
R2 (Spatial Proximity)    & 65.68 & 0.58 & 58.50 & 0.51 & 70.36 & 0.60 \\
R3 (Local Environment) & 65.70 & 0.57 & 58.65 & 0.50 & 70.44 & 0.60 \\
\bottomrule
\end{tabular}
\end{adjustbox}
\end{center}
\end{table*}

{Table \ref{tab:ablation_results_2} presents an ablation study assessing the impact of different relational factors—Sequential Relationship (R1), Spatial Proximity (R2), and Local Environment (R3)—on three-state predictions across CB513, CASP12, and TS115 datasets. The results demonstrate that each relation contributes comparably to the overall performance, with minor variations in accuracy and F1 scores. Notably, the full SSRGNet model does not significantly outperform the individual relations, suggesting that each relation independently captures meaningful structural information. However, the small differences highlight the complementary nature of these relational cues, justifying their inclusion in the model. The study underscores the necessity of these three relations in constructing an effective graph representation.}

\section{Conclusion}
In conclusion, this study addresses the challenging task of predicting secondary structures from protein primary sequences, a crucial step towards understanding tertiary structures and unraveling insights into protein functions and relationships. Existing methods often rely solely on unlabeled amino acid sequences, neglecting the valuable 3D structural data that profoundly influences protein functionality. To bridge this gap, we introduce protein residue graphs and incorporate various forms of sequential and structural connections to capture richer spatial information.

Our approach combines Graph Neural Networks (GNNs) and Language Models (LMs). We utilize a pre-trained transformer-based protein language model to encode amino acid sequences and leverage message-passing mechanisms like GCN and R-GCN to capture the geometric properties of protein structures. By applying convolution within a node's local region and considering relationships, we stack multiple convolutional layers to efficiently extract insights from the protein's spatial graph, uncovering intricate interconnections and dependencies in its structural arrangement.

For evaluating our model's performance, we use the NetSurfP-2.0 training dataset, which provides secondary structure annotations in both 3- and 8-states. Our extensive experiments demonstrate that our proposed model, called SSRGNet, achieves impressive F1-scores of 61.32\%, 51.22\%, and 65.45\% on publicly available test datasets, specifically CB513, TS115, and CASP12, as evaluated by the Q3 metric. This highlights the effectiveness of our approach in accurately predicting protein secondary structures and suggests its potential utility in various protein-related applications.

{In the future, we aim to address the several limitations identified in our current work. First, we plan to integrate external structural datasets to pre-train the structure-based encoders, such as R-GCN or transformer-based models, using contrastive learning. This approach can enable the encoders to learn a more diverse and robust set of structural patterns, improving generalization even in cases of sparse or imperfect 3D data. Additionally, pretraining on protein sequences can help these encoders better capture long-range dependencies that are otherwise difficult to model through local message passing alone. We also aim to refine the current parallel fusion strategy by introducing adaptive gating or attention-based mechanisms to dynamically weigh the contributions from both sequence and structural modalities.}

\section{Ethical Considerations}
We make use of publicly available datasets. Without violating any copyright issues, we followed the policies of the datasets we used.

\end{document}